%% file: acl_latex.tex
\pdfoutput=1

\documentclass[11pt]{article}

\usepackage[final]{acl}

\usepackage{times}
\usepackage{latexsym}

\usepackage[T1]{fontenc}

\usepackage[utf8]{inputenc}

\usepackage{microtype}

\usepackage{inconsolata}

\usepackage{graphicx}
\usepackage{subcaption}

\usepackage{booktabs}
\usepackage{amssymb,amsmath,amsthm}
\usepackage{bm}
\usepackage{pifont}
\usepackage{multirow}
\newcommand{\model}{LCIRC}

\title{LCIRC: A Recurrent Compression Approach for Efficient Long-form Context and Query Dependent Modeling in LLMs}

\author{
Sumin An\textsuperscript{1} \qquad
Junyoung Sung\textsuperscript{1} \qquad
Wonpyo Park\textsuperscript{2}
\\
\textbf{Chanjun Park\textsuperscript{1,\(\dagger\)}} \qquad
\textbf{Paul Hongsuck Seo\textsuperscript{1,\(\dagger\)}}
\\
\textsuperscript{1}Dept. of CSE, Korea University
\qquad
\textsuperscript{2}Google
\\
\texttt{\{suminan, jys7451, bcj1210, phseo\}@korea.ac.kr}
\\
\texttt{wppark@google.com}
}

\begin{document}
\maketitle
\begin{abstract}
While large language models (LLMs) excel in generating coherent and contextually rich outputs, their capacity to efficiently handle long-form contexts is limited by fixed-length position embeddings. Additionally, the computational cost of processing long sequences increases quadratically, making it challenging to extend context length. 
To address these challenges, we propose Long-form Context Injection with Recurrent Compression (LCIRC), a method that enables the efficient processing long-form sequences beyond the model's length limit through recurrent compression without retraining the entire model.
We further introduce query dependent context modeling, which selectively compresses query-relevant information, ensuring that the model retains the most pertinent content. Our empirical results demonstrate that Query Dependent LCIRC (QD-LCIRC) significantly improves LLM's ability to manage extended contexts, making it well-suited for tasks that require both comprehensive context understanding and query relevance.

\end{abstract}
\let\thefootnote\relax\footnotetext{$^\dagger$Co-corresponding authors.}

\input{sections/introduction}
\input{sections/related_work}
\input{sections/method_final}
\input{sections/experiments_final}
\input{sections/conclusion}

\section*{Limitations}
Although our proposed method, LCIRC, demonstrates notable improvements in handling long-form contexts, several limitations remain. First, the training and evaluation of query dependent context modeling were restricted to question-answering tasks, leaving its generalizability to other query-driven applications, such as information retrieval and dialogue systems, unexplored. Future work is needed to assess its broader applicability. Furthermore, despite the reduction in computational complexity during inference, LCIRC incurs substantial training costs due to the introduction of cross-attention layers, which increases both the computational load and the amount of training data required. This constraint may limit the method’s deployment in resource-constrained environments. Lastly, the experimental evaluation was primarily conducted on English-language datasets, limiting the conclusions that can be drawn about LCIRC performance in multilingual or non-English contexts. Further experimentation on diverse linguistic datasets is necessary to evaluate the model’s cross-lingual capabilities.

\section*{Ethics Statement}
All experiments conducted in this study were performed with fairness and transparency in mind. The datasets used were sourced from publicly available data, and no personally identifiable information (PII) was involved. Our experimental design ensured that all models were trained and evaluated under consistent conditions, allowing for fair comparisons. We commit to maintaining high ethical standards in the use and development of our models, promoting responsible research practices.

\section*{Acknowledgments}
This research was supported by the IITP grants (RS-2020-II201819, RS-2024-00436857, RS-2024-00398115), the NRF grants (NRF-2021R1A6A1A03045425, RS-2023-00280034) and the KOCCA grant (RS-2024-00345025) funded by the Korea government (MSIT, MOE and MSCT).

\bibliography{custom}


\end{document}

%% file: sections/introduction.tex
\section{Introduction}

Large language models (LLMs) have demonstrated remarkable capabilities in generating coherent and contextually rich outputs~\cite{zhao2023survey,wang2024beyond}. However, a major limitation remains in their ability to handle long-form contexts efficiently~\cite{liu2024lost,li2024long}. Transformer-based architectures, which underlie most LLMs, are constrained by fixed-length position embeddings, limiting the number of tokens they can process in a single pass~\cite{chen2021simple,lin2022survey}. Once this limit is exceeded, crucial information from earlier parts of the input is truncated, leading to reduced performance in tasks requiring extended context, such as document summarization~\cite{koh2022empirical} or long-form question answering~\cite{fan2019eli5}. Also, the computational cost of processing longer input sequences grows quadratically with sequence length, making it impractical to simply increase the context window~\cite{fournier2023practical,chen2023extending}.

To address these challenges, we propose Long-form Context Injection with Recurrent Compression (\model), a method designed to extend LLM's ability to process long-form inputs efficiently. LCIRC compresses the context beyond the model's length limit into compact representations, which are injected back into the model, allowing it to retain essential information without retraining the entire model. By recurrently compressing the input sequence, our approach maintains a balance between preserving context and minimizing computational overhead, enabling the model to generate outputs grounded in long-span contexts~\cite{valmeekam2023llmzip,chen2024melodi,lester2024training}.

In many real-world applications, LLMs need to process input in response to specific queries or instructions~\cite{ouyang2022training,naveed2023comprehensive}. Compressing all available information indiscriminately can lead to irrelevant data being retained, reducing the model’s effectiveness~\cite{mulc2024compressing,franceschelli2024training}. To address this, we introduce query dependent context modeling, which selectively compresses information based on its relevance to the query. This ensures that the model focuses on the most pertinent parts of the input, enhancing its performance in tasks like multiple choice and long-form question answering, where query relevance is crucial.

Our results show that \model, combined with query dependent modeling, significantly improves the ability of LLMs to handle long-form contexts in a scalable and efficient manner, making it well-suited for applications that require both extensive context understanding and precise query relevance.

In summary, our primary contributions are as follows: (1) We propose LCIRC, a method that extends the context window of LLMs through recurrent compression, allowing efficient handling of long-form inputs without retraining the entire model. (2) We introduce query dependent context modeling, which selectively compresses query-relevant information, enhancing performance in tasks that require the comprehensive understanding of extended contexts. (3) Our approach significantly improves the efficiency of LLMs in long-form contexts, reducing computational costs while demonstrating performance improvements quantitatively through various benchmarks.



%% file: sections/related_work.tex
\section{Related Work}

\begin{figure*}[t]
    \centering  
    \includegraphics[width=0.95\linewidth]{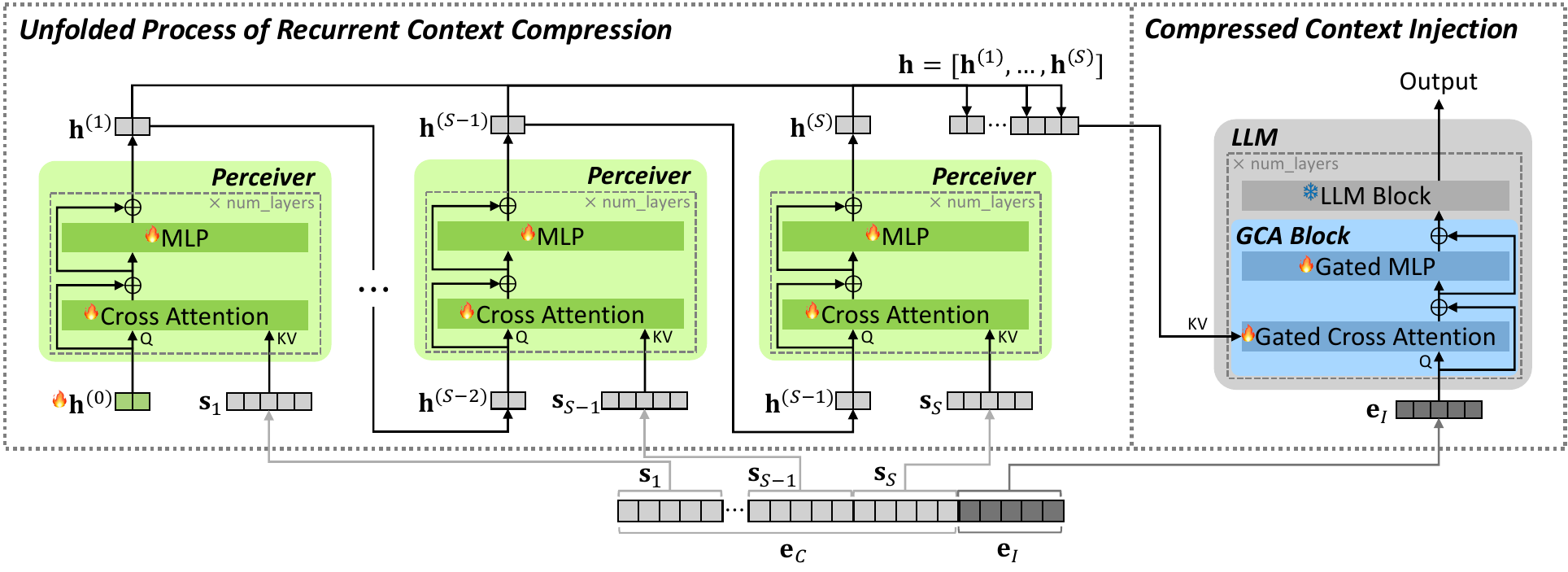}
    \vspace{-0.3cm}
    \caption{\textbf{The overall process of the proposed Long-form Context Injection with Recurrent Compression (LCIRC)}. LCIRC comprises two components: Recurrent Context Compression (left) and Compressed Context Injection (right). In the $i$-th step of Recurrent Context Compression, the previously compressed features $\mathbf{h}^{(i-1)}$ and the segment embeddings $\mathbf{s}_{i}$ are fed into the Perceiver module as query and input features, respectively. The compressed features $\mathbf{h}^{(i)}$ are then generated and reinjected as query features for the subsequent recurrence step. The initial query features $\mathbf{h}^{(0)}$ are learnable parameters. In Compressed Context Injection, the concatenated compressed features $\mathbf{h}$ serve as input to the Gated Cross Attention layer. Layers indicated with a fire symbol represent trained layers, while layers marked with a snow symbol denote frozen layers.}
    \vspace{-0.3cm}

    \label{fig:overview}
\end{figure*}

\paragraph{Recurrent Transformers}
Various methods have been explored to extend the capacity of transformer-based language models for handling longer input sequences through the use of recurrence mechanisms. These approaches typically divide the input into segments and recurrently store and reuse information across them. Transformer-XL~\cite{transformer-xl} and Compressive Transformer~\cite{compressive} employ segment-level recurrence, while models such as Recurrent Memory Transformer~\cite{rmt} and Memformer~\cite{memformer} utilize memory modules to retain past information. Additionally, Recurrent Attention Network (RAN)~\cite{RAN} and Segmented Recurrent Transformer (SRformer)~\cite{SRformer} integrate recurrent attention to aggregate information across segments. Despite these advancements, a common limitation of these methods is the need for full model re-pretraining, which is computationally costly and resource-intensive.

\paragraph{Sparse Attention}
As input sequence lengths increase, the computational overhead associated with full attention mechanisms grows quadratically. Sparse attention mechanisms have been proposed as an efficient alternative to mitigate this computational burden. Notable models such as Sparse Transformer~\cite{sparse-transformer}, Longformer~\cite{longformer}, Reformer~\cite{reformer}, Big Bird~\cite{bigbird}, Routing Transformer~\cite{routing-transformer}, and MInference~\cite{minference} implement sparse attention to reduce computational costs for longer contexts.
While these methods improve efficiency, they do so at the cost of partial information loss, resulting in performance that often falls short of full attention mechanisms, and they are still constrained by the limited context length of LLMs, making it challenging to process long-form sequences.

\paragraph{Prompt Compression}
Prompt compression techniques aim to reduce the computational challenges of processing large prompts by creating more compact representations. Gisting~\cite{gist} compresses the entire user instruction into a set of "gist tokens" in one step, providing a compact representation of the prompt. However, this approach is typically optimized for shorter contexts, limiting its effectiveness when handling inputs that exceed the context window of the underlying LLM.
Additionally, several studies have explored 
KV cache compression~\cite{dmc, pyramidkv} to enhance the efficiency of long-form context inference. 
While these approaches improve computational efficiency, they are primarily designed to optimize context modeling within the context length of the underlying LLM and have limited capability in extending the sequence length.
ICAE~\cite{icae} and AutoCompressor~\cite{autocompressor} address this limitation by segmenting long contexts into smaller chunks and compressing each individually. Although these methods represent progress, they have yet to demonstrate compression capabilities for extremely long sequences, such as those involving hundreds of thousands of tokens, and still face limitations in terms of inference costs and context window size due to the simultaneous inference and compression processes.

%% file: sections/method_final.tex
\section{Efficient Long-form Context Modeling}
In this section, we present an efficient approach to handle long-form inputs in pretrained LLMs. We introduce a method that enables models to process lengthy contexts by compressing and injecting the relevant information back into the model in a computationally efficient manner. Additionally, we outline the training strategy used to optimize the proposed components while maintaining the foundational capabilities of the model.

\newcommand{\vx}{\mathbf{x}}
\newcommand{\vy}{\mathbf{y}}
\newcommand{\ve}{\mathbf{e}}
\newcommand{\vh}{\mathbf{h}}
\newcommand{\vs}{\mathbf{s}}

\subsection{Preliminaries: Transformer-Based LLMs}
Modern LLMs are built on the transformer architecture, where the model autoregressively estimates the probability of each subsequent token $x_{n+1:N}$ given the preceding tokens $x_{1:n}$, as formulated by:
\begin{align}
    P(x_{n+1:N}|x_{1:n})=\prod_{i=n+1}^{N} P(x_{i}|x_{1:i-1})
\end{align}
Here, $x_{j:k}$ refers to a subsequence of tokens $(x_j, x_{j+1}, \dots, x_{k})$, and $x_{1:N}$ denotes the full token sequence. During generation, the model samples one token at a time, appending it to the input to predict the next token.

Accurately modeling token order is essential in autoregressive generation because it enables the model to effectively capture the underlying structure and meaning of language. Transformer-based LLMs employ learned positional embeddings to represent token positions, but these embeddings impose a fixed input length, limiting the model’s ability to handle sequences longer than the maximum number of embeddings ($M$). Consequently, transformers are constrained in their capacity to process input sequences exceeding this limit ($N > M$).

Incorporating long-form context into LLMs is critical for generating outputs that remain grounded in extended input sequences. However, extending the input length requires full retraining of the model, and the computational cost scales quadratically with the context length, making long-form input processing computationally prohibitive in standard transformer architectures.

\subsection{Long-form Context Injection with Recurrent Compression}
To address the limitations of processing lengthy inputs ($N \gg M$) in pretrained LLMs, we propose Long-form Context Injection with Recurrent Compression (\model{}), an approach that enables efficient handling of long-form contexts. A simple truncation of the first $N-M$ tokens would discard essential contextual information, whereas our method restores access to this context through recurrent context compression and compressed context injection, detailed below. An overview of the proposed method is provided in Figure~\ref{fig:overview}.

\subsubsection{Recurrent Context Compression}
\label{subsubsection:RCC}
We introduce a recurrent context compression mechanism that effectively reduces the long-form context into a compact sequence of embeddings, which can be efficiently processed by the model.

Given an input sequence $x_{1:N}$, where $x_{1:N-M}$ represents the truncated context $x_C$, it is often the case that $N \gg M$, for example, $N = 192\mathrm{K}$ in InfiniteBench~\cite{infinitebench} compared to $M = 4\mathrm{K}$ in Llama~\cite{llama2}. To handle this, the recurrent compressor produces a compact feature sequence $(h_1, \dots, h_K) \in \vh$ from $x_C$, where $K \ll N-M$.

We employ the Perceiver architecture~\cite{perceiver, perceiverio} for the compressor, which consists of stacked Perceiver blocks. Each block includes a cross-attention layer followed by a two-layer MLP with residual connections (Figure~\ref{fig:overview}). The cross-attention mechanism aggregates input features based on query features, enabling efficient compression by using a compact sequence as the query and the longer context as the input features.

In particular, we use the token embeddings $\ve_C$ of the truncated context $x_C$ as the input features and a set of learnable query vectors $\vh^{(0)}$ of length $K$ as the query features. The compressed features $\vh$ are obtained through the Perceiver module as follows:
\begin{align}
    \vh = \mathrm{Perceiver}(\vh^{(0)}, \ve_C)
\end{align}
where $\mathrm{Perceiver}(q, x)$ represents the Perceiver module with query $q$ and input features $x$.

However, compressing such an extensive context in a single step is computationally expensive, thus we introduce a recurrent compression process. The long context $\ve_C$ is split into $S$ disjoint segments $\vs_1, \dots, \vs_S$, where $\vs_i = \ve_{n_{i-1}+1:n_i}$ represents the $i$-th segment. These segments are sequentially fed into the Perceiver module, with the compressed features from the previous segment serving as the query features for the next segment:
\begin{align}
    \vh^{(i)} = \mathrm{Perceiver}(\vh^{(i-1)}, \vs_i)
    \label{eq:vanillaRCC}
\end{align}
Here, $\vh^{(i)}$ compresses both the current segment $\vs_i$ and the cumulative information from all previous segments, enabled by the recurrent mechanism. The initial query vectors $\vh^{(0)}$ consists of learnable parameters.

Finally, the compressed representations $\vh^{(1)}, \dots, \vh^{(S)}$ of all segments are concatenated to form the overall compressed representation of the long-form context:
\begin{align}
    \vh = [\vh^{(1)}, \dots, \vh^{(S)}]
\end{align}
where $[\cdots]$ denotes concatenation. This recurrent approach ensures efficient long-form context representation, enabling LLMs to process extended inputs beyond their native length limitations.

\subsubsection{Compressed Context Injection}
\label{subsubsection:CCI}
After obtaining the compressed representation $\vh$ for the long-form context, we inject this compressed information into the pretrained transformer using gated cross-attention layers with residual connections~\cite{flamingo}. For the truncated input sequence $x_{N-M+1:N}$ of length $M$, denoted as $x_{I}$
,we first compute the embedding sequence $\ve^{(l)}_{I}$ at the $l$-th transformer block. The embeddings are then contextualized through Gated Cross Attention Block (GCA Block in Figure \ref{fig:overview}) as follows:
\begin{align}
\begin{split}
    \Dot{\ve}^{(l)}_{I} &= \alpha^{(l)} \cdot \mathrm{CA}(\ve^{(l)}_{I}, \vh) + \ve^{(l)}_{I} \\
    \alpha^{(l)} &= \mathrm{tanh}(a^{(l)}) \\
    \Ddot{\ve}^{(l)}_{I} &= \beta^{(l)} \cdot \mathrm{MLP}(\Dot{\ve}^{(l)}_{I}) + \Dot{\ve}^{(l)}_{I} \\
    \beta^{(l)} &= \mathrm{tanh}(b^{(l)})
\end{split}
\label{eq:GCAB}
\end{align}
where $\mathrm{CA}(q, x)$ denotes the cross-attention layer with queries $q$ and key-value inputs $x$. Unlike standard transformer layers, we pass the modified embeddings $\Ddot{\ve}^{(l)}_{I}$ to the next transformer block instead of the original embeddings $\ve^{(l)}_{I}$. The scalar parameters $a^{(l)}$ and $b^{(l)}$ are learnable and initialized to 0, preserving the pretrained LLM’s performance at the start of training. The compressed context representation enables efficient context injection, minimizing computational overhead.

\subsubsection{Optimization Strategy}
\label{subsubsection:optimization}
To fully leverage the pretrained LLM's foundational capabilities, we optimize only the additional components using a corpus of long-form texts.

Given a long-form training sequence $x_{1:N}$, the model is trained by minimizing the negative log-likelihood (NLL) loss:
\begin{align}
    \mathcal{L} &= -\frac{1}{N} \sum_{i=1}^{N} \log P(x_i | x_{1:i-1}).
    \label{eq:training_objective}
\end{align}
The long-form input is randomly segmented, with each segment limited to a maximum length $R$. The probability $P(x_i | x_{1:i-1})$ is estimated within each segment, and prior segments are processed by the recurrent context compressor. The input $x_{k:i-1}$, where $k$ is the starting index of the current segment, is treated as the regular input for the LLM.

Although the recurrent architecture enables memory-efficient inference, the space complexity during training scales linearly with the input length $N$ due to backpropagation through time (BPTT). To mitigate this, we employ truncated BPTT, where gradient computation is restricted to the last $T$ segments. Since gradient calculation is unnecessary for earlier segments beyond the last $T$, we cache the compressed features $\vh$ and reuse them for predicting subsequent tokens within each segment.

\subsubsection{Inference}
\label{sec:inference}
LLMs perform inference by conditioning on a given input prompt and generating a coherent and contextually relevant output sequence.
Given an input token sequence of length $N$ and an output token sequence of maximum length $P$, if their combined length remains within the context window $M$ of the underlying LLM ($N+P \le M$), the our recurrent compression mechanism is not activated, as LCIRC is specifically designed for long-form contexts.
In this case, the model operates in the same manner as the inference process of the pretrained LLMs.
In contrast, when the total maximum length of input and output sequences exceeds the context window of the LLM ($N+P > M$), two distinct cases must be considered.
When the maximum output length remains within the context window of the LLM ($P \le M$), the LLM can generate the entire output sequence in a single pass given the compressed input sequence by the proposed recurrent compression mechanism.
Conversely, if the maximum output length exceeds the context window ($P > M$), our model iteratively compresses the earlier portions of the generated tokens (e.g., the first ${M}/{2}$ tokens) by performing additional recurrent steps using the compression mechanism.
This process removes the compressed tokens from the context of the underlying LLM, thereby creating space for generating new next tokens while preserving coherence and continuity. 
Notably, this additional process remains highly efficient, as it only involves the compression steps while maintaining the original window size of the underlying LLM.

\section{Query Dependent Context Modeling}
LLMs frequently process context based on specific instructions or queries. To enhance the ability of LLMs to handle long-form context, we extend our method to incorporate query dependent context compression. This allows the model to selectively focus on the context most relevant to the given query, thus improving the overall efficiency and relevance of the model's responses.

\begin{figure}[t]
    \centering  
    \begin{subfigure}{.47\linewidth}
        \centering
        \includegraphics[width=1\linewidth]{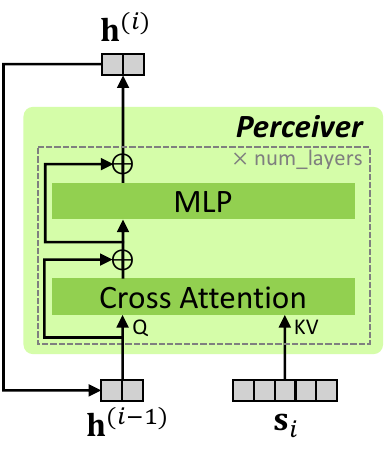}
    \end{subfigure}\hspace{0.3cm}
    \begin{subfigure}{.47\linewidth}
        \centering
        \includegraphics[width=1\linewidth]{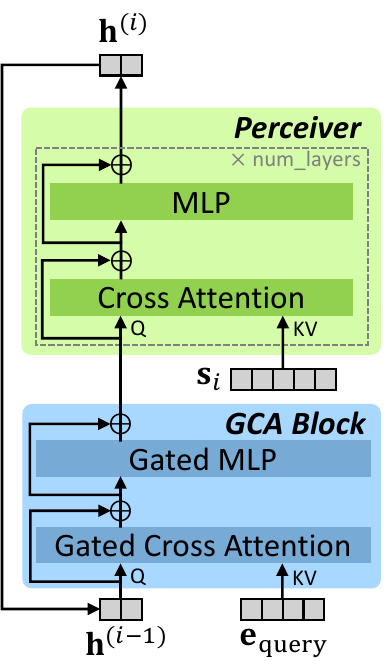}
    \end{subfigure}
    \caption{\textbf{Comparison of the recurrent context compression module with and without query dependent modeling.} 
    In addition to the regular context compression module (left), we add additional cross attention module (blue box) to inject query information into the compressed feature $\vh^{(i-1)}$ (right).}
    \label{fig:QDRCC}
\end{figure}

Unlike the vanilla recurrent context compression, which merges all the information in the current segment $\vs_i$ into the compressed features $\vh^{(i)}$ as shown in Eq.~\eqref{eq:vanillaRCC}, query dependent compression selectively injects information that is most relevant to the user query. This selective compression is achieved through the addition of a Gated Cross Attention Block to our recurrent context compression method as shown in Figure~\ref{fig:QDRCC}.

\begin{figure*}[t]
    \centering  
    \includegraphics[width=\linewidth]{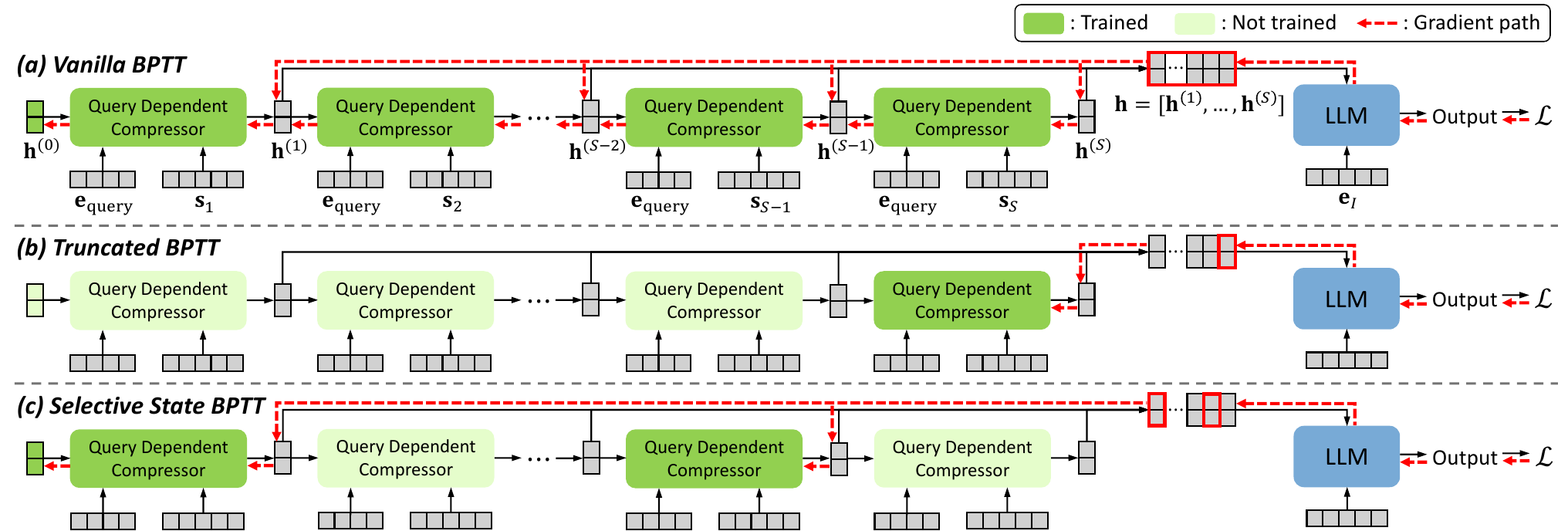}
    \caption{
    \textbf{Comparisons of the proposed Selective State BPTT with vanilla and truncated BPTT.}
    Green boxes represent timesteps where gradients are computed in BPTT whereas the light green ones indicate the timesteps without gradient computation. 
    Finally, dotted red lines illustrate the gradient flows.
    (a) Vanilla BPTT computes the full gradients through the entire timesteps in recurrence but is computationally infeasible with a large $N$. The gradients for $\vh^{(i)}$ receives upstream gradients both through the recurrent connection and through the direct connection from $\vh$. 
    (b) Truncated BPTT backprobagates gradients to the last $T$ timesteps only significantly reducing computational costs.
    However, it does not transfer gradient flows to timesteps further than $T$ (marked with light green color) and fails to learn long-term QD modeling.
    (c) Our proposed Selective State BPTT selects several random timesteps and transfer gradient flows directly through the direct connection from $\vh$, which enables efficient learning of long-term QD modeling capabilities.
    }
    \label{fig:bptt}
\end{figure*}

As illustrated in Figure~\ref{fig:QDRCC}, the model integrates the user query into the compression pipeline. At each compression step, the learnable query vectors or the previously compressed features $\vh^{(i-1)}$ used as the query features for the Perceiver module are transformed into a query dependent representation. This is done through the newly introduced gated cross attention block, which takes $\vh^{(i-1)}$ as the query features, and the user query embedding $\ve_{\mathrm{query}}$ as the input features. The query dependent features $\Ddot{\vh}^{(i-1)}$ are computed through the same process in Eq.~\eqref{eq:GCAB} with $\vh^{(i-1)}$ and $\ve_{\mathrm{query}}$.

The query dependent compressed feature $\vh^{(i)}$ is then produced using the Perceiver module, with the query dependent feature $\Ddot{\vh}^{(i-1)}$ and the segment embeddings $\vs_i$ through the same process in Eq.~\eqref{eq:vanillaRCC}.
Subsequently, these query dependent compressed features $\vh$ are then used during inference as described in Eq.~\eqref{eq:GCAB}, enabling the model to focus on information relevant to the query while handling long-form inputs.

\paragraph{Training and Efficiency Enhancements}
Query Dependent LCIRC (QD-LCIRC) builds upon the pre-trained LCIRC architecture by adding a gated cross-attention layer to introduce query dependency. To train QD-LCIRC, we minimize the following negative log-likelihood (NLL) loss:
\begin{align}
    \mathcal{L} &= -\frac{1}{N} \sum_{i=1}^{N} \log P(x_i | x_{1:i-1}, x_{\mathrm{query}})
\end{align}
This objective ensures that the model learns to predict each token in the input sequence conditioned on both the preceding tokens and the query, facilitating query dependent context modeling.

As shown in Figure~\ref{fig:bptt}, we employ Random Selective BPTT to train the QD-LCIRC efficiently. Unlike vanilla BPTT, which computes gradients across all timesteps, or truncated BPTT, which only computes gradients for the last $T$ timesteps, Random Selective BPTT randomly selects a subset of timesteps for gradient computation. This allows the model to efficiently learn long-term query dependent context modeling without excessive computational overhead. Additionally, we cache the compressed features $\vh$ from earlier segments, further reducing the memory and computational requirements for training.

\paragraph{Inference}
The only distinction between QD-LCIRC and LCIRC lies in the incorporation of query dependent compression, facilitated by the GCA block, within the recurrent compression step. Consequently, the inference process of QD-LCIRC differs from that of LCIRC only by the addition of query dependent modeling through an extra GCA block in the recurrent compression step, as demonstrated in Section~\ref{sec:inference} and Figure~\ref{fig:QDRCC}.


%% file: sections/experiments_final.tex
\section{Experiments}

\subsection{Datasets and Metrics}
\begin{table}[h]
\small
\centering
\begin{tabular}{rrr}
\toprule
\textbf{Token Length} & \textbf{\# of Samples} & \textbf{Proportion (\%)} \\ \midrule
6K $\leq$ Data $<$ 8K & 410,876 & 36.64 \\
8K $\leq$ Data $<$ 16K & 418,332  & 37.30 \\
16K $\leq$ Data $<$ 32K & 207,503 & 18.50 \\
32K $\leq$ Data $<$ 64K & 69,396 & 6.19 \\
64K $\leq$ Data $<$ 128K & 13,659 & 1.22 \\
128K $\leq$ Data & 1,679 & 0.15 \\ \midrule
\textbf{Total} & 1,121,445 & 100.00 \\ \bottomrule
\end{tabular}
\caption{
\textbf{Data distribution of the training set extracted from FineWeb-Edu.} Our training set is constructed focusing on long-context modeling.
}
\label{tab:data_stats}
\end{table}

\paragraph{FineWeb-Edu}
FineWeb-Edu~\cite{finewebedu} comprises 1.3T tokens, filtered from the FineWeb dataset, which contains 15T tokens. To facilitate long-form context modeling, we selected texts with a minimum length of 4K tokens, resulting in a dataset that includes texts up to 339K tokens in length. The statistics of the training data are provided in the Table~\ref{tab:data_stats}. For evaluation, we curated 1,000 texts of 128K tokens each to assess perplexity.

\paragraph{FineWeb-LQA}
FineWeb-LQA, a long-form QA dataset, was automatically generated from FineWeb-Edu to support the training of our query-dependent model, following a data construction process similar to \citet{in2}. We extract random 128-token text segments and utilize the Llama-3.1-70B-Instruct-FP8 model to generate corresponding QA pairs. The extracted segments are then reintegrated into the original long-form context, which serves as the basis for evaluating long-form QA tasks.

\paragraph{InfiniteBench}
InfiniteBench~\cite{infinitebench} is a benchmark suite designed to assess the capability of LLMs in handling ultra-long contexts, exceeding 100K tokens. We focus on two tasks: LongBook QA (En.QA) and LongBook Multiple Choice (En.MC), both of which test the model's ability to answer identical questions in open-ended and multiple-choice formats, respectively. Evaluation is conducted using the F1 score for En.QA and accuracy for En.MC.

\paragraph{LongBench}
LongBench~\cite{longbench} evaluates long-form context modeling through a suite of tasks across real-world and synthetic categories. In this study, we focus on six English tasks, consisting of both single-document and multi-document QA tasks. The average context length is 18K tokens, with a maximum of 82K tokens. Performance is measured using the F1 score.

\paragraph{L-Eval}
L-Eval~\cite{leval} includes a dataset of 508 long documents spanning various domains, divided into 20 subtasks. It comprises over 2K human-annotated query-response pairs, with context lengths ranging from 3K to 200K tokens. We evaluate the models on four open-ended QA tasks using the F1 score and four multiple-choice QA tasks using accuracy.

\subsection{Models}
We build upon Llama2-7B~\cite{llama2} as our baseline model, augmenting its long-form context modeling capabilities. In line with \cite{autocompressor}, we implement an extended version of Llama2 that supports full attention over longer contexts (ExtendedFA), extending the token length limit to 8K by modifying the RoPE $\theta$~\cite{rope}. Due to the quadratic complexity of full attention, ExtendedFA is restricted to a maximum of 8K tokens. We compare this approach against AutoCompressor~\cite{autocompressor}, a state-of-the-art method that models context through prompt compression by recurrently feeding segmented inputs to the model while leveraging compressed tokens for subsequent segments. Additionally, we evaluate our proposed method with and without query dependent modeling.


All models are trained on FineWeb-Edu to ensure fair comparisons in language modeling. For QD-LCIRC, we initialize the model with the pre-trained weights of LCIRC on FineWeb-Edu and fine-tune it on FineWeb-LQA for query dependent modeling. 
Note that FineWeb-LQA is automatically generated from FineWeb-Edu for this purpose.

For baseline models, context exceeding the token length limit is truncated. Llama2 and ExtendedFA handle up to 4K and 8K tokens, respectively, while AutoCompressor supports up to 84K tokens, based on the experimental setup from \citet{autocompressor}. In contrast, our method imposes no explicit length limit, as the compressed context is injected via cross-attention, allowing us to process sequences up to 815K tokens in length.

\subsection{Implementation Details}
All models are based on Llama2-7B and trained using a batch size of 64 with the Adam optimizer \cite{adam}.
QD-LCIRC is trained with a learning rate of 2e-5 and 300 warmup steps,
utilizing Selective State BPTT with 8 random selections.
Other models are trained with a learning rate of 5e-5.
The length ($K$) of the initial learnable queries $\vh^{(0)}$, which also corresponds to the length of each compressed features $\vh^{(i)}$, is set to 64 based on our preliminary experiments with a smaller model OPT-2.7B.
In these experiments, we observed no significant performance difference between $K=64$ and $K=256$, leading us to adopt the more efficient setting of $K=64$.
All training procedures are conducted using eight NVIDIA H100 80GB GPUs.

\begin{table}[t]
\small
\centering
\begin{tabular}{lcccc}
\toprule
 & \multicolumn{4}{c}{Total Token Length ($N$)}\\
 \cmidrule(lr){2-5}
Models & 4k & 8k & 64k & 128k\\ \midrule
Llama-2-7B & 5.472 & - & - & - \\
ExtendedFA & \textbf{5.442} & 5.319 & - & - \\
AutoCompressor & 6.127 & 6.010 & 6.188 & -\\
LCIRC (Ours) & 5.472 & 5.313 & 5.312 & 5.312\\
QD-LCIRC (Ours) & 5.472 & \textbf{5.299} & \textbf{5.298} & \textbf{5.298}\\ \bottomrule
\end{tabular}
\caption{
\textbf{Perplexity scores on the FineWeb-Edu test set.} 
Each long-form text is truncated from the beginning to adhere to the total token length, which encompasses both the context and the last 2K target tokens used for measuring perplexity.
}
\label{table:ppl}
\end{table}

\begin{table}[t]
\small
\centering
\begin{tabular}{lcccc}
\toprule
 & \multicolumn{4}{c}{Total Token Length ($N$)}\\
 \cmidrule(lr){2-5}
Models & 4k & 8k & 64k & 128k\\ \midrule
ExtendedFA & 63 & 143 & 3,118 & 10,739 \\
AutoCompressor & 61 & 125 & 1,350 & -\\
LCIRC (Ours) & 63 & 77 & 97 & 120\\
QD-LCIRC (Ours) & 63 & 77 & 98 & 122\\ \bottomrule
\end{tabular}
\caption{
\textbf{Computational complexities for different models in TeraFLOPs.}
We compute the TFLOPs of ExtendedFA under the assumption that the model is extended to process input tokens of the specified length.
AutoCompressor is unable to process inputs with 128K tokens.
}
\label{table:cost}
\end{table}

\begin{table*}[t]
\centering
\begingroup
\setlength{\heavyrulewidth}{0.10em}  
\setlength{\lightrulewidth}{0.055em}
\resizebox{\textwidth}{!}{
\begin{tabular}{l cc ccc ccccccc}
\toprule
 & & & \multicolumn{3}{c}{\textbf{InfiniteBench}} & \multicolumn{7}{c}{\textbf{LongBench}}\\
 \cmidrule(lr){4-6}  \cmidrule(lr){7-13}
 & \textbf{FW-LQA} & \textbf{QD} & \textbf{En.MC} & \textbf{En.QA} & \textbf{Avg} & \textbf{NQA} & \textbf{Qasper} & \textbf{MFQA} & \textbf{HQA} & \textbf{2WQA} & \textbf{MSQ} & \textbf{Avg}\\ \midrule
Llama-2-7B & \ding{55} & \ding{55} & 6.99 & 3.95 & 5.47 & \textbf{13.04} & 12.08 & 14.68 & 16.27 & 7.10 & 4.41 & 11.26\\
ExtendedFA & \ding{55} & \ding{55} & 15.72 & 3.88 & 9.80 & 12.21 & 18.23 & 18.23 & 17.81 & 14.18 & 8.25 & 14.82\\
AutoCompressor & \ding{55} & \ding{55} & 18.34 & 4.46 & 11.40 & 12.60 & 16.89 & 19.93 & 19.00 & 16.36 & 8.84 & 15.60\\
LCIRC (Ours) & \ding{55} & \ding{55} & \textbf{21.40} & \textbf{5.26} & \textbf{13.33} & 10.67 & \textbf{18.32} & \textbf{21.71} & \textbf{21.66} & \textbf{16.55} & \textbf{9.09} & \textbf{16.33}\\ \midrule
ExtendedFA & \ding{51} & \ding{55} & 28.38 & 4.55 & 16.47 & \textbf{18.96} & 13.73 & 23.48 & 20.78 & 17.24 & 8.29 & 17.08\\
AutoCompressor & \ding{51} & \ding{55} & 31.00 & 5.35 & 18.18 & 13.69 & 18.63 & \textbf{33.55} & 15.01 & 14.13 & 8.98 & 17.33\\
QD-LCIRC (Ours) & \ding{51} & \ding{51} & \textbf{38.86} & \textbf{5.80} & \textbf{22.33} & 15.31 & \textbf{20.57} & 33.25 & \textbf{28.19} & \textbf{19.00} & \textbf{12.39} & \textbf{21.45}\\ \bottomrule
\end{tabular}
}
\endgroup
\caption{\textbf{Per-task performance on InfiniteBench and LongBench.} The following abbreviations are used: \textbf{NQA} denotes NarrativeQA, \textbf{MFQA} represents MultiFieldQA-en, \textbf{HQA} refers to HotpotQA, \textbf{2WQA} to 2WikiMQA, and \textbf{MSQ} to MuSiQue. \textbf{Avg} indicates the average score across all subtasks within respective benchmarks.
\textbf{FW-LQA} indicates whether the model is fine-tuned on FineWeb-LQA.
Our QD-LCIRC consistently outperforms competing methods, achieving the highest average score by incorporating query dependent modeling, as indicated in the \textbf{QD} column.
}
\label{tab:infinite_long_bench}

\end{table*}

\begin{table*}[t]
\small
\centering
\begingroup
\setlength{\heavyrulewidth}{0.10em}  
\setlength{\lightrulewidth}{0.055em}
\resizebox{\textwidth}{!}{
\begin{tabular}{l ccccccccccc}
\toprule
 & \textbf{FW-LQA} & \textbf{QD} & \textbf{CS} & \textbf{QALIT} & \textbf{TOEFL} & \textbf{SF} & \textbf{LFQA} & \textbf{NQA} & \textbf{NQ} & \textbf{Qasper} & \textbf{Avg} \\ \midrule
Llama-2-7B & \ding{55} & \ding{55} & 10.47 & 25.74 & 0.00 & 29.69 & 22.30 & 8.66 & 22.64 & 2.29 & 15.22\\
ExtendedFA & \ding{55} & \ding{55} & 16.86 & 33.66 & 17.10 & 22.66 & 13.89 & 10.76 & 33.24 & 10.96 & 19.89\\
AutoCompressor & \ding{55} & \ding{55} & 18.60 & 31.68 & 17.47 & 22.66 & 11.73 & 4.82 & 26.42 & 6.64 & 17.50 \\
LCIRC (Ours) & \ding{55} & \ding{55} & 22.09 & 27.23 & 10.04 & 27.34 & 14.66 & 9.89 & 26.40 & 9.44 & 18.39\\ \midrule
ExtendedFA & \ding{51} & \ding{55} & 15.70 & \textbf{34.16} & 13.75 & 20.31 & 28.27 & 16.91 & \textbf{35.68} & 8.51 & 21.66 \\
AutoCompressor & \ding{51} & \ding{55} & 20.35 & 32.18 & \textbf{26.39} & 16.41 & 29.33 & 19.08 & 28.67 & 15.38 & 23.47 \\
QD-LCIRC (Ours) & \ding{51} & \ding{51} & \textbf{25.58} & 30.20 & 12.27 & \textbf{37.50} & \textbf{31.63} & \textbf{20.92} & 34.37 & \textbf{16.92} & \textbf{26.17}\\ \bottomrule
\end{tabular}
}
\endgroup
\caption{\textbf{Per-task performance on L-Eval}. The following abbreviations are used: \textbf{CS} denotes Coursera, \textbf{QALIT} refers to QuALITY, \textbf{SF} represents SFiction, \textbf{LFQA} refers to LongFQA, and \textbf{NQA} to NarrativeQA. \textbf{Avg} indicates the mean performance score across all subtasks within the respective benchmark. \textbf{FW-LQA} indicates whether the model has been fine-tuned on FineWeb-LQA, while \textbf{QD} denotes whether query dependent modeling.}
\label{tab:leval}
\end{table*}

\begin{table}[t]
\small
\centering
\begin{tabular}{lcccc}
\toprule
 & InfBench & LongBench & L-Eval \\ \midrule
Truncated BPTT & 21.26 & 20.73 & 25.45\\
Selective State BPTT & \textbf{22.33} & \textbf{21.45} & \textbf{26.17}\\ \bottomrule
\end{tabular}
\caption{\textbf{Impact of different BPTT variations on benchmark performance.} This table presents the performance of the model trained with Truncated BPTT, and Selective State BPTT ($T=8$) across three benchmarks: InfiniteBench, LongBench, and L-Eval. The results highlight the differences in benchmark scores for each method, demonstrating the effectiveness of Selective State BPTT.}
\label{table:cos}
\end{table}
\vspace{-0.2cm}

\subsection{Results on Language Modeling}
Following~\citet{autocompressor}, we first measure the perplexity of the last 2K tokens given its context while varying the total token length between 4K and 128K, which refers to the combined length of both the context and target tokens.
To ensure that the same tokens are used for perplexity measurement for comparisons, we truncate the inputs from the beginning based on the total context length ($N$), retaining the last $N$ tokens, of which the final 2K tokens are used for the perplexity computation.
Note that all the models except ours impose limits on context lengths. 

Table~\ref{table:ppl} presents the perplexity results of the models on the FineWeb-Edu test set.
We first observe that all the methods, except for AutoCompressor, perform similarly when $N=4$K.
AutoCompressor significantly alters Llama's generation mechanism resulting in a notable drop in perplexity.
This finding aligns with the observations from experiments conducted with Llama in \citet{autocompressor}.
In contrast, our method preserves the original strong capabilities of Llama with short contexts, as the frozen LLM remains unchanged.

With $N>4$K, Llama is unable to process the full context.
In contrast, all other models exhibit improved perplexity scores by utilizing additional context compared to their scores with $N=4$K. 
Note, however, that both ExtendedFA and AutoCompressor still impose strict limits on the total context length. 
Moreover, the perplexity of AutoCompressor increases at $N=64$K, indicating challenging optimization for long-form context modeling. 
In contrast, the proposed LCIRC model consistently improves perplexity and maintains this improved performance even as the context further lengthens.

We evaluate the computational complexities of the models for processing long-term contexts with varying total token lengths at inference, as shown in Table~\ref{table:cost}. 
As the token length increases, the results show a prohibitively large increase in complexity with ExtendedFA, which requires full attention across all tokens.
In contrast, AutoCompressor effectively reduces complexity compared to ExtendedFA; 
a reduction rate increases with the number of tokens and showing 66\% reduction with 64K tokens. 
However, AutoCompressor cannot process tokens longer than 64K with segments of 2K token length.
In contrast, LCIRC can handle 128K tokens, while achieving a 99\% complexity reduction compared to ExtendedFA. 
Note that LCIRC improved perplexity while maintaining this significant reduction in complexity.
Finally, QD-LCIRC introduces some additional complexities, but they are marginal compared to the overall reduction rate.

\subsection{Results on Long Context Benchmarks}
We also evaluate the methods on multiple QA benchmarks that require long-form context understanding.
In these benchmarks, models are asked to answer a question, requiring to understand the input text under the context of the question. 
For QD-LCIRC, we use the input question as the query.

Table~\ref{tab:infinite_long_bench} presents performances on InfiniteBench and LongBench.
Since the QA instances in these benchmarks require an understanding of long-form context documents, Llama exhibits poor performance across all tasks.
ExtendedFA improves this by leveraging additional context, but it is still limited to 8K tokens, sharing the same underlying issue as Llama.
Both AutoCompressor and LCIRC further enhance performance over ExtendedFA by enabling access to much longer contexts, with LCIRC achieving much greater improvements.
When LCIRC is combined with our query-dependent modeling technique (QD-LCIRC), it leads to significant performance gains, yielding approximately 308\% and 90\% relative improvements in average scores over the base Llama model on InfiniteBench and LongBench, respectively.
To ensure fair comparisons, we further fine-tuned ExtendedFA and AutoCompressor on FineWeb-LQA.
Despite this, our QD-LCIRC still achieves significant improvements over these models, consistently delivering the best performance on most tasks and resulting in the highest average score.
This further underscores the effectiveness of incorporating query dependent modeling.


We also evaluate the models on L-Eval in Table~\ref{tab:leval}.
Note that, although L-Eval is a benchmark designed to assess long-form context understanding capabilities, its context lengths are relatively shorter, with an average length of 19K compared to 218K in InfiniteBench. 
Based on this, Llama, which can process up to 4K tokens, demonstrates strong performance on several tasks. 
When extended to capture longer contexts using various methods—namely ExtendedFA, AutoCompressor, and LCIRC—all models show improved average performance compared to the base model.
Finally, our QD-LCIRC achieves the highest average performance on L-Eval as well, demonstrating significant gains through query dependent modeling, with a relative improvement of approximately 11.5\% compared to the best-performing baseline, ExtendedFA finetuned on FineWeb-LQA.

In Table~\ref{table:cos}, we compare the proposed selective state BPTT with truncated BPTT for QD-LCIRC on InfiniteBench, LongBench and L-Eval.
The results show that our selective state BPTT allows higher scores compared to truncated BPTT across all three benchmarks.
Note that truncated BPTT only backpropagates gradients to a limited number of timesteps in recurrence, restricting optimization for long-term context modeling.
In contrast, our selective state BPTT enables the model to receive gradients from any timesteps and in consequence, the trained model better models long inputs.

%% file: sections/conclusion.tex
\section{Conclusion}
We propose Long-form Context Injection with Recurrent Compression (LCIRC) to address challenges LLMs face with extended inputs. LCIRC efficiently compresses long-form contexts, expanding context length while reducing computational overhead. By incorporating query dependent context modeling, it selectively retains relevant information, improving performance in tasks requiring long-context comprehension and query relevance. Our experiments demonstrate significant advancements in both areas. Future work will focus on extending LCIRC to multilingual settings to integrate context across diverse languages and cultures.